\documentclass[11pt]{article}

\usepackage[final]{acl}

\usepackage{times}
\usepackage{latexsym}
\usepackage{booktabs}
\usepackage{amsfonts}
\usepackage{subcaption}
\usepackage{listings}
\usepackage{xcolor}
\lstdefinestyle{promptstyle}{
  basicstyle=\ttfamily\small,
  backgroundcolor=\color{gray!10},
  frame=single,
  framerule=0pt,
  breaklines=true,
  columns=fullflexible,
  keepspaces=true,
  xleftmargin=1em,
  xrightmargin=1em
}
\usepackage{tabularx}
\usepackage{xcolor}
\usepackage{tcolorbox}
\usepackage{multirow}
\usepackage{enumitem}
\usepackage{float}

\usepackage{amsmath} 
\usepackage[T1]{fontenc}
\usepackage[utf8]{inputenc}

\usepackage{microtype}

\usepackage{inconsolata}

\usepackage{graphicx}
\usepackage{color}
\usepackage{array}

\title{ActivityEditor: Learning to Synthesize Physically Valid Human Mobility}

\author{
  Chengjie Yang\textsuperscript{1}, 
  Yutian Jiang\textsuperscript{2}, 
  Anqi Liang\textsuperscript{3}, 
  Wei Qi\textsuperscript{4}, 
  \textbf{Chenyu Wu}\textsuperscript{1,$\ast$}, 
  \textbf{Junbo Zhang}\textsuperscript{5,$\dagger$} \\
  \textsuperscript{1}Southwest Jiaotong University,
  \textsuperscript{2}HKUST (Guangzhou),
  \textsuperscript{3}Shanghai Jiao Tong University, \\
  \textsuperscript{4}Tsinghua University,
  \textsuperscript{5}JD Technology \\
  \texttt{3335028104@my.swjtu.edu.cn}, \texttt{yjiang194@connect.hkust-gz.edu.cn}, \\
  \texttt{\{anqiliang9657, chenywu60\}@gmail.com}, \\
  \texttt{qiw@tsinghua.edu.cn}, \texttt{msjunbozhang@outlook.com}
  \thanks{~~$\ast$Lead corresponding author. $\dagger$Corresponding author.}
}

\begin{document}
\maketitle
\begin{abstract}
Human mobility modeling is indispensable for diverse urban applications. However, existing data-driven methods often suffer from data scarcity, limiting their applicability in regions where historical trajectories are unavailable or restricted. To bridge this gap, we propose \textbf{ActivityEditor}, a novel dual-LLM-agent framework designed for zero-shot cross-regional trajectory generation. 
Our framework decomposes the complex synthesis task into two collaborative stages. Specifically, an intention-based agent, which leverages demographic-driven priors to generate structured human intentions and coarse activity chains to ensure high-level socio-semantic coherence. These outputs are then refined by editor agent to obtain mobility trajectories through iteratively revisions that enforces human mobility law. This capability is acquired through reinforcement learning with multiple rewards grounded in real-world physical constraints, allowing the agent to internalize mobility regularities and ensure high-fidelity trajectory generation.  Extensive experiments demonstrate that \textbf{ActivityEditor} achieves superior zero-shot performance when transferred across diverse urban contexts. It maintains high statistical fidelity and physical validity, providing a robust and highly generalizable solution for mobility simulation in data-scarce scenarios. Our code is available at: \url{https://anonymous.4open.science/r/ActivityEditor-066B}.
\end{abstract}

\section{Introduction}
Human mobility modeling is a cornerstone of modern urban intelligence, providing critical insights for a wide spectrum of real-world applications, including map services \cite{ye2011exploiting}, logistics dispatching \cite{wen2024survey}, urban planning \cite{zheng2023spatial}, and public safety monitoring \cite{gao2017identifying}. Despite its significance, the efficacy of these applications is heavily predicated on the availability of high-fidelity trajectory data. In practice, obtaining such data is increasingly challenging due to data scarcity arising from factors such as distribution shifts in evolving urban environments\cite{feng2020learning} and the ``cold-start'' problem in newly planned areas where historical records have yet to be generated\cite{he2020human}. Coupled with stringent privacy regulations\cite{gong2025stcdm}, this vacuum of mobility data has become a significant bottleneck, hindering the deployment of intelligent urban services.

In recent years, deep learning-based models \cite{long2023practical, wang2023synthesizing, yuan2022activity} have been widely applied to mobility synthesis, achieving promising results by capturing high-order transition dynamics and mining shared mobility patterns among users. However, these data-driven approaches suffer from several key drawbacks. First, their success heavily relies on the collection of large-scale private mobility data, which is often inaccessible due to stringent privacy regulations. Second, such models are inherently data-hungry and struggle to generalize in zero-shot settings, where historical trajectories for a target region are entirely absent. Finally, the prediction accuracy remains limited by the constrained sequential modeling capabilities of traditional architectures and a conspicuous lack of deep understanding regarding human daily commonsense. 

While recent advancements in Large Language Models (LLMs) have shown promise in preliminary mobility prediction\cite{wang2023would,feng2025agentmove} and synthesis\cite{shao2024chain,du2025cams}, they continue to face critical bottlenecks. Specifically, most existing LLM-based methods still require partial local trajectories to perceive regional patterns, hindering their application in strict zero-shot scenarios. Furthermore, these methods typically rely on single-pass generation architectures that lack self-correction or auditing mechanisms, often leading to spatio-temporal inconsistencies and violations of physical laws in real-world environments.

We propose ActivityEditor, a dual-agent framework designed for high-fidelity, zero-shot mobility synthesis. Our approach is grounded in the insight that while urban functional indicators and population-level metrics vary across cities, the underlying mapping between socio-demographic characteristics and human intentions remains invariant(Evidence can be seen in Appendix \ref{appendix:dataset}). ActivityEditor takes individual profiles as input and decomposes the synthesis process into two collaborative stages. In the Intention-driven Generation stage, an agent leverages demographic priors to derive structured intentions and construct a coarse activity skeleton. This is followed by Rule-constrained Refinement, where an Editor agent iteratively enforces human mobility laws via "Add, Delete, Shift, and Replace" operations to establish logical consistency and rectify spatio-temporal ungrounding. To ensure the reliability of this refinement, we utilize Group Relative Policy Optimization (GRPO) to align the agent with real-world physical regularities. This enables the model to autonomously internalize physical laws and correct latent cognitive biases regarding human movement, allowing ActivityEditor to perform robust zero-shot generation in entirely unseen target cities by learning a universal logic of trajectory refinement.

Our contributions are summarized as follows:
\begin{itemize}
    \item We define and address the challenge of zero-shot human mobility synthesis, enabling the generation of realistic trajectories starting solely from socio-demographic profiles. This approach effectively bypasses the dependency on private historical data and provides a scalable solution for the ``cold-start'' problem in newly planned urban areas.
    
    \item We propose ActivityEditor, a novel dual-LLM-agent framework that treats mobility synthesis as an iterative ``Generate-then-Edit'' task. By decomposing the process into intention-driven generation and rule-constrained refinement, and further employing GRPO for physical-law alignment, we establish a self-correcting mechanism that ensures both semantic coherence and spatio-temporal validity.

    \item  Extensive experiments across diverse urban scenarios demonstrate that ActivityEditor significantly outperforms state-of-the-art baselines. Our results confirm the model's superior zero-shot generalizability, achieving remarkable improvements in physical consistency and statistical fidelity without requiring any local trajectory data for the target regions.
\end{itemize}

\section{Related Work}
\subsection{Mobility Modeling} Existing research can be categorized into mobility prediction and mobility synthesis. Early methods in both fields primarily utilized deep learning architectures, such as RNNs and GANs, to capture spatiotemporal dynamics \cite{gao2019predicting, feng2018deepmove, song2016deeptransport, feng2020learning}. However, these statistical approaches often struggle to incorporate high-level semantic knowledge, leading to suboptimal performance in complex urban scenarios. To bridge this gap, recent studies have integrated Large Language Models (LLMs) to enhance predictive capabilities \cite{li2024large, wang2023would, zhong2025comapoi, feng2024wheretomove} and generative fidelity \cite{wang2024cola, shao2024chain}.Meanwhile, emerging LLM-based mobility generation frameworks \cite{wang2024large,liu2024human,liu2026gatsim, wu2024mrpllm} have further explored agent-centric, profile-conditioned, and city-aware simulation paradigms, improving the realism and personalization of synthetic travel diaries and activity chains \cite{amin2025generating,li2024geo}.

Despite their semantic depth, these methods exhibit significant limitations in zero-shot generalization and physical consistency. Most LLM-based models still exhibit sensitivity to local data distributions or require few-shot samples to perceive regional patterns \cite{feng2025agentmove, han2025unimove, li2025rallmpoi}. Crucially, current paradigms are predominantly single-pass (forward-only) architectures that lack self-correction or auditing mechanisms. This often leads to spatio-temporal inconsistencies, such as violations of travel velocity constraints or logical gaps in activity chains. Unlike these approaches, ActivityEditor treats mobility modeling as an iterative "generate-then-edit" task, aligning synthetic trajectories with physical laws and human commonsense through a dual-agent refinement process.

\subsection{LLM and Agentic Reinforcement Learning}  Post-training alignment has evolved from Reinforcement Learning from Human Feedback (RLHF) \cite{ouyang2022training} to more efficient paradigms like Direct Preference Optimization (DPO) \cite{rafailov2023direct}. Recently, Group Relative Policy Optimization (GRPO) \cite{guo2025deepseek} and its variants, such as DAPO \cite{yu2025dapo}, have emerged to enhance complex reasoning by normalizing rewards within sample groups without explicit value functions,and have further inspired multi-turn agent reinforcement learning with finer-grained turn-level optimization and reward design \cite{zeng2025reinforcing,wei2025reinforcing}. Transitioning toward Agentic RL, research frames LLMs as autonomous decision-makers that optimize long-horizon trajectories through environment feedback \cite{zhang2025landscape,zhou2025sweet, qi2025webrl}. Unlike standard alignment, agentic approaches emphasize iterative self-correction and tool-use reasoning\cite{feng2025retool, wang2025tool4poi, dong2025toolstar}. ActivityEditor advances this by integrating GRPO-based alignment into a dual-agent "generate-then-edit" framework, enabling the autonomous internalization of physical regularities for zero-shot mobility synthesis.

\section{Preliminary}
We define the zero-shot mobility synthesis task and related concepts for use in the following section.

\textbf{Socio-demographic Attributes:} 
For each agent $i$, we define their background as a set of $K$ attributes, denoted by $D_i = \{d_{1,i}, d_{2,i}, \dots, d_{K,i}\}$. These attributes encompass both individual-level demographics (e.g., age, gender, employment) and household-level descriptors (e.g., income, household size, vehicle ownership), serving as the initial prompt seeds for mobility generation.

\textbf{Human Intention:} 
Distinct from prior works, we introduce the concept of latent intention to bridge demographics and behaviors. Let $I_{n,i} \in \mathcal{V}_I$ be the underlying motivation (e.g., \textit{sustenance}, \textit{education}, \textit{recreation}) for the $n$-th activity of agent $i$. An intention chain is defined as a time-ordered sequence: $\mathcal{I}_i = \{I_{1,i}, I_{2,i}, \dots, I_{n,i}\}$.

\textbf{Activity and Activity Chain:} 
An individual activity $A_{n,i}$ is defined as a tuple $[T_{n,i}, S_{n,i}, E_{n,i}]$, where $T_{n,i}$ denotes the activity type (semantic category), while $S_{n,i}$ and $E_{n,i}$ represent the start and end times, respectively. An activity chain $A_i = \{A_{1,i}, A_{2,i}, \dots, A_{n,i}\}$ represents the complete sequence of events for agent $i$ over a 24-hour period.

\textbf{Mobility Trajectory:} 
The final synthetic trajectory $Traj_i$ is expressed as a sequence of activity-location pairs: $Traj_i = \{(A_{1,i}, Z_{1,i}), \dots, (A_{n,i}, Z_{n,i})\}$, where $Z_{n,i}$ denotes the zone-level location (e.g., Transportation Analysis Zone, TAZ) where the activity occurs.

\textbf{Zero-shot Mobility Synthesis:} 
Given a set of socio-demographic attributes $D_i$ from an unseen target region $\mathcal{R}_{test}$, the objective is to learn a generative mapping $\mathcal{M}: D_i \to Traj_i$. The model $\mathcal{M}$ must achieve high statistical fidelity and physical validity without having observed any historical trajectory samples from $\mathcal{R}_{test}$ during the training phase.

\section{Methods}

\subsection{Overview}
\begin{figure*}[t] 
\centering 
\includegraphics[width=\textwidth]{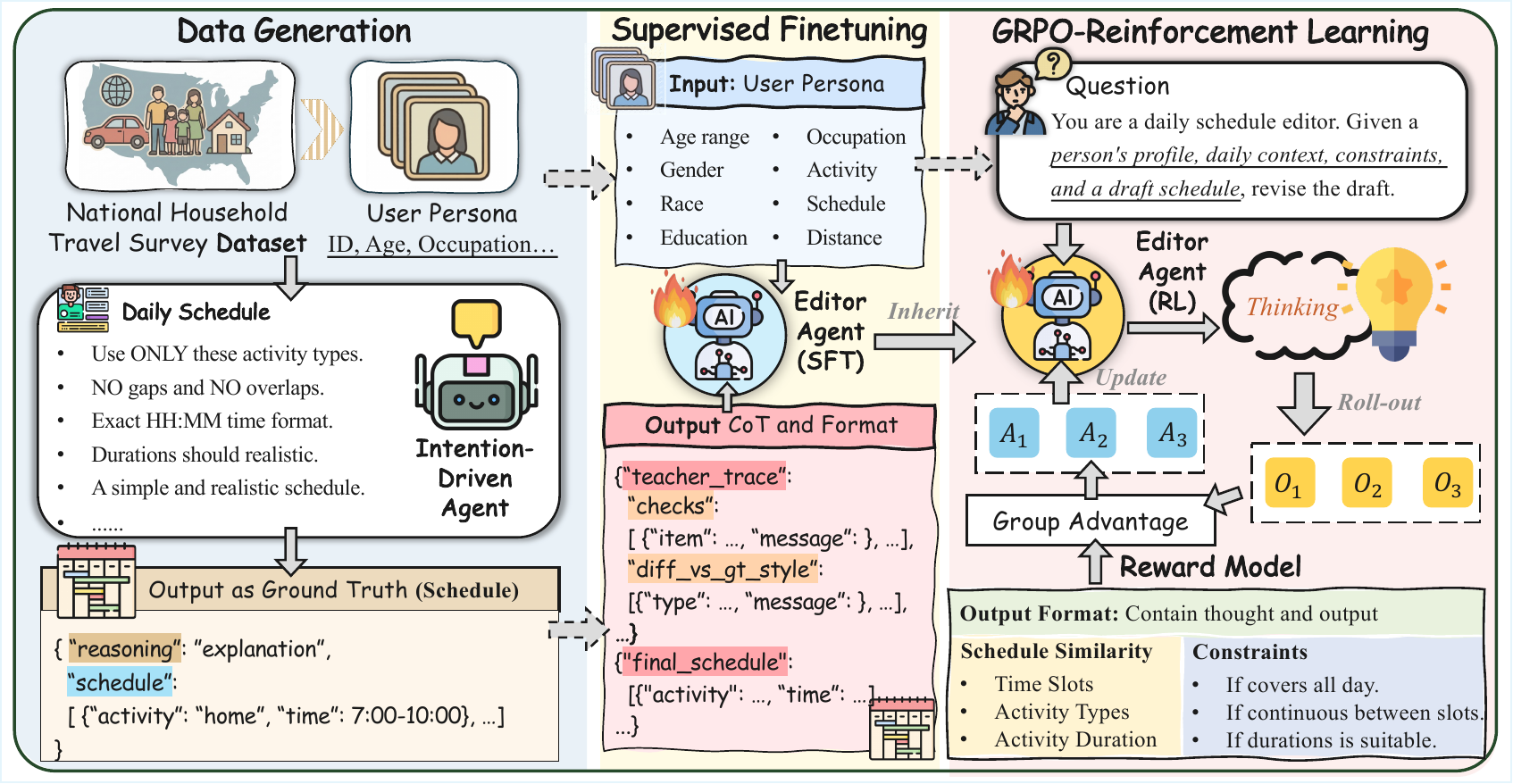} 
\caption{Overview of the \textsc{ActivityEditor} framework.} 
\label{fig:overview} 
\vspace{-0.5em} 
\end{figure*}

As illustrated in Figure \ref{fig:overview}, \textsc{ActivityEditor} comprises two core modules: the Intention-Driven Agent and the Editor Agent. Serving as the semantic architect, the Intention-Driven Agent is designed to induce the Socio-Semantic Prior by mapping demographic profiles into a activity draft. To rectify inherent Mobility Biases and instill an explicit awareness of spatio-temporal regularities, the Editor Agent functions as a refinement module that performs Add, Delete, Shift, and Replace operations. While the Intention-Driven Agent operates in a zero-shot manner to preserve universal behavioral priors, the Editor Agent is specifically optimized through a two-stage training pipeline involving SFT and GRPO to ensure the final synthesis achieves both socio-semantic coherence and strict physical validity. Detailed prompt templates and task-specific instructions for both agents are provided in Appendix \ref{appendix:prompts}.

\subsection{Intention-driven Agent}
The Intention-driven Agent induces the Socio-Semantic Prior, defined as the cross-regionally similar mapping between demographics and behavioral intentions. While urban functional indicators vary, the underlying logic linking user profiles to intentions remains consistent. By leveraging a frozen LLM as a world model, the agent infers latent intentions directly from user profiles to generate foundational human mobility patterns.

\paragraph{Input Normalization.}
The agent receives a structured user profile $D_i$, comprising $K$ diverse attributes such as \texttt{age\_range}, \texttt{employment\_status}, \texttt{relationship}, and \texttt{primary\_activity}.

\paragraph{Skeleton Generation.}
Upon receiving $D_i$, the agent initiates a reasoning trace within the thought space to infer the user's latent intention, denoted as $V_I$. By leveraging its parametric knowledge of population clustering and behavioral patterns, the agent articulates the rationale behind the user's daily demands. For example, if a user profile indicates \texttt{primary\_activity = ``Retired''}, the agent may explicitly reason that ``the user is retired and therefore does not require a daily work commute.''Guided by these inferred intentions, the agent constructs a preliminary draft schedule, denoted as $S_{\text{draft}}$. This draft establishes the chronological order of the activity chain and allocates rough temporal intervals.

% \paragraph{Semantic Coherence.}
% It is important to note that the primary objective of this stage is to ensure semantic coherence and statistical fidelity with respect to the demographic input. Consequently, $S_{\text{draft}}$ serves as an ``activity sequence skeleton.'' Although it captures the foundational macroscopic mobility patterns, it is allowed to contain minor temporal imperfections, such as imprecise durations or slight time-step overlaps, which then serve as the raw material for the subsequent editor agent to refine and rectify.

\subsection{Editor Agent}
The Editor Agent acts as a rule-based optimizer to bridge the gap between abstract intentions and physical reality. While the Intention-driven Agent ensures semantic logic, the resulting $S_{\text{draft}}$ may fail to satisfy the underlying spatio-temporal constraints of human mobility. The Editor Agent specifically injects explicit mobility regularities to rectify the spatio-temporal ungrounding inherent in LLMs. By functioning as a rule-aware auditor, it transforms these coarse drafts into strictly valid trajectories by enforcing rigorous physical and logical boundaries.

\paragraph{Constraint Checking.}
The agent first acts as a rigorous ``critic'' within the \texttt{<Thought>} space, comprehensively auditing the draft to identify any violations of predefined rules. As illustrated in the ``Reward Model'' module of Figure~\ref{fig:overview}, these constraints are formalized into two major categories:

\textbf{1) Hard Constraints.} These are strictly enforced rules that ensure the fundamental mathematical and structural validity of the trajectory:
\begin{itemize}[topsep=2pt, itemsep=2pt, parsep=0pt]
    \item \textit{Physical Constraints:} Absolute spatiotemporal boundaries. For example, the activity chain seamlessly cover a complete 24-hour period.
    \item \textit{Logical Constraints:} Inherent structural and state-transition mandates. Consecutive segments in the schedule must represent distinct states; transitioning directly from one activity to the exact same activity type is structurally invalid and must be merged into a single continuous segment.
\end{itemize}

\textbf{2) Soft Constraints.} These act as heuristic guidelines to evaluate the behavioral realism, socioeconomic alignment, and profile consistency of the sequence:
\begin{itemize}[topsep=2pt, itemsep=1pt, parsep=0pt]
    \item \textit{Commonsense Constraints:} Alignment with socio-demographic profiles. For example, a user explicitly profiled as \textit{retired} should not be assigned regular, prolonged \textit{work} activities.
    \item \textit{Temporal Constraints:} Plausible duration boundaries for specific activities. For example, while sleeping or working can naturally take several hours, a \textit{drop-off/pick-up} or \textit{service} event should be relatively brief and cannot unrealistically span half a day.
    \item \textit{Coherence Constraints:} Macroscopic sequence rationality. This prevents excessive temporal fragmentation, such as alternating between home and shopping every 10 minutes, thereby ensuring smooth and logical activity transitions.
\end{itemize}

\paragraph{Action and Iterative Refinement.}
Upon detecting any of the aforementioned violations, the agent triggers an internal ReAct loop\cite{yao2022react}. It selects specific editing operations from a predefined action space to locally modify the draft. These operations include \textsc{Add} (inserting missing activities), \textsc{Delete} (removing irrational events), \textsc{Shift} (adjusting start/end times to resolve overlaps), \textsc{Replace} (swapping contextually inappropriate activities), and \textsc{Split} (dividing excessively long activities proportionally). After each execution, the agent re-evaluates the updated schedule until all conflicts are resolved. Once it confirms that the schedule satisfies all spatiotemporal and physical rules, the agent outputs the finalized, high-fidelity JSON-structured trajectory.

\subsection{Training Strategy}
Our training pipeline, comprising SFT and GRPO, is designed to systematically eliminate the latent Mobility Bias and hallucinations of pre-trained LLMs. This process replaces noisy, sparse internal representations with verified domain knowledge, allowing the Editor Agent to internalize the fundamental laws of human movement. Through this two-stage alignment, the model shifts from generic reasoning to a precise, rule-governed cognitive model of mobility.

\paragraph{Data Generation.}
Real-world human mobility datasets, such as travel surveys, exclusively record the final executed trajectories. They inherently lack the intermediate cognitive processes of planning, constraint checking, and conflict resolution. To bridge this gap and construct a training corpus for the editing mechanism, we employ a powerful teacher model, GPT-5.2, to perform a reverse-engineering task. Specifically, we supply the teacher model with the user demographic profile, the flawed preliminary draft generated by the Intention-driven Agent, and the corresponding ground-truth schedule. Tasked with mapping the draft to the ground truth, the teacher model infers a highly detailed Chain-of-Thought trace. This trace explicitly articulates the step-by-step logic required to identify temporal overlaps or commonsense violations in the draft, along with the precise refinement operations needed to correct them. Consequently, we successfully synthesize the missing cognitive traces, creating high-quality paired data of reasoning processes and structural outputs.

% \vspace{-5pt}
\paragraph{Supervised Fine-Tuning.}
Utilizing the synthesized reasoning dataset, we perform supervised fine-tuning to initialize the student model as the Editor Agent. This phase is fundamentally designed to solve the cold-start problem inherent in complex generation tasks. By training strictly on the paired thought traces and the final JSON formats, the student model acquires foundational domain knowledge regarding human mobility patterns. More importantly, it learns the specific structural paradigm of self-critique. The model masters how to properly open a thought space, sequentially propose editing actions such as adding, deleting, or shifting activities, and ultimately output a rigidly formatted JSON trajectory.

% \vspace{-5pt}
\paragraph{GRPO-Reinforcement Learning.}
To transcend mere data imitation and deeply internalize physical laws, we introduce GRPO in the second phase. Given a user profile, the policy model parallelly performs roll-outs to generate a group of candidate schedules. Subsequently, a programmatic, rule-based reward model evaluates these outputs without human-in-the-loop intervention. The composite reward is formulated through three explicit dimensions:

\textbf{ Formatting Reward ($R_{\text{fmt}}$):} This ensures adherence to the ``Generate-then-Edit'' paradigm. It assigns scores based on the presence of the reasoning process (\texttt{[THOUGHT]}) and the structured output (\texttt{[JSON]}):
\begin{equation}
R_{\text{fmt}} = 
\begin{cases} 
1.0, & \text{if both \texttt{[THOUGHT]} and \texttt{[JSON]} exist} \\ 
0.5, & \text{if only one tag exists} \\ 
0.0, & \text{otherwise} 
\end{cases}
\end{equation}

\textbf{ Constraint Reward ($R_{\text{con}}$):} This validates the physical feasibility of the trajectory. Let $\mathcal{S} = \{(s_1, e_1), (s_2, e_2), \dots, (s_N, e_N)\}$ denote the generated schedule containing $N$ segments, where $s_i$ and $e_i$ are the start and end times in minutes. $R_{\text{con}}$ is the average of three Boolean checks:
\begin{equation}
R_{\text{con}} = \frac{1}{3} \big( \mathbb{I}_{\text{full}} + \mathbb{I}_{\text{cont}} + \mathbb{I}_{\text{dur}} \big)
\end{equation}
where $\mathbb{I}_{\text{full}}$ ensures full-day coverage; $\mathbb{I}_{\text{cont}}$ enforces temporal continuity without gaps or overlaps; and $\mathbb{I}_{\text{dur}}$ constrains each activity duration to a realistic physiological range.

\textbf{ Fidelity Reward ($R_{\text{sim}}$):} This evaluates the statistical realism of the generated sequence compared to standard mobility patterns. It combines the token-level accuracy (Acc), Macro-F1 score, and the Jensen-Shannon Divergence (JSD) of activity type distributions ($p_{\text{act}}$) and interval length distributions ($p_{\text{int}}$):
\begin{equation}
\begin{aligned}
R_{\text{sim}} &= 0.40 \cdot \text{Acc} + 0.10 \cdot \text{F1}_{\text{macro}} \\
&\quad + 0.25 \cdot \big(1 - \text{JSD}(p_{\text{gen}}^{\text{act}} \parallel p_{\text{gt}}^{\text{act}})\big) \\
&\quad + 0.25 \cdot \big(1 - \text{JSD}(p_{\text{gen}}^{\text{int}} \parallel p_{\text{gt}}^{\text{int}})\big)
\end{aligned}
\end{equation}
where $p_{\text{gen}}$ and $p_{\text{gt}}$ denote the probability distributions of the generated schedule and the ground truth, respectively.

Based on the sum of these rewards $R = R_{\text{fmt}} + R_{\text{con}} + R_{\text{sim}}$, the group advantage is calculated to update the policy model. Through this group-relative comparison, model autonomously learns to increase generation probability of high-quality CoTs, thereby thoroughly internalizing spatiotemporal constraints and self-editing capabilities.

\section{Experiment}

\subsection{Experimental Setup}

\begin{table*}[h]
\centering
\small
\renewcommand{\arraystretch}{1.1} 
\setlength{\tabcolsep}{1pt}
\resizebox{\textwidth}{!}{
\begin{tabular}{l ccc ccc ccc ccc}
\toprule
\multicolumn{1}{c}{\multirow{2}{*}{\textbf{Method}}} & \multicolumn{3}{c}{\textbf{California}} & \multicolumn{3}{c}{\textbf{Arizona}} & \multicolumn{3}{c}{\textbf{Georgia}} & \multicolumn{3}{c}{\textbf{Oklahoma}} \\
\cmidrule(lr){2-4} \cmidrule(lr){5-7} \cmidrule(lr){8-10} \cmidrule(lr){11-13}
& \textit{Acc} $\uparrow$ & \textit{Mint} $\downarrow$ & \textit{Atype} $\downarrow$ & \textit{Acc} $\uparrow$ & \textit{Mint} $\downarrow$ & \textit{Atype} $\downarrow$ & \textit{Acc} $\uparrow$ & \textit{Mint} $\downarrow$ & \textit{Atype} $\downarrow$ & \textit{Acc} $\uparrow$ & \textit{Mint} $\downarrow$ & \textit{Atype} $\downarrow$ \\
\midrule

% --- Traditional Baselines ---
\multicolumn{13}{l}{{Traditional Baselines}} \\
\quad\quad\quad\quad $\diamond$ DeepMove & 0.554 & 0.660 & 0.229 & 0.521 & 0.666 & 0.252 & 0.448 & 0.657 & 0.264 & 0.472 & 0.676 & 0.281 \\
\quad\quad\quad\quad $\diamond$ LSTPM    & 0.402 & 0.705 & 0.290 & 0.420 & 0.691 & 0.300 & 0.379 & 0.693 & 0.288 & 0.376 & 0.697 & 0.287 \\
\cmidrule(lr){1-13}

% --- General LLM Baselines ---
\multicolumn{13}{l}{{General LLM Baselines}} \\
\quad\quad\quad\quad $\diamond$ ChatGPT & 0.660 & 0.425 & 0.199 & 0.664 & 0.399 & 0.195 & 0.583 & 0.403 & 0.201 & 0.616 & 0.404 & 0.158 \\
\quad\quad\quad\quad $\diamond$ Qwen    & 0.553 & 0.535 & 0.228 & 0.551 & 0.598 & 0.194 & 0.511 & 0.583 & 0.238 & 0.542 & 0.570 & 0.167 \\
\quad\quad\quad\quad $\diamond$ LLaMA   & 0.526 & 0.576 & 0.291 & 0.524 & 0.643 & 0.248 & 0.486 & 0.624 & 0.305 & 0.515 & 0.611 & 0.213 \\
\cmidrule(lr){1-13}

% --- Specific LLM ---
\multicolumn{13}{l}{{Specific LLM}} \\
\quad\quad\quad\quad $\diamond$ CoPB    & 0.703 & 0.577 & \underline{0.141} & 0.674 & 0.579 & \underline{0.157} & 0.602 & 0.603 & \textbf{0.152} & 0.627 & 0.567 & 0.140 \\
\quad\quad\quad\quad $\diamond$ RAGHome & 0.679 & 0.556 & 0.183 & 0.658 & 0.548 & 0.198 & 0.593 & 0.574 & 0.167 & 0.614 & 0.528 & 0.142 \\
\cmidrule(lr){1-13}

% --- Fine-tuned LLMs ---
\multicolumn{13}{l}{{Fine-tuned LLMs}} \\
\quad\quad\quad\quad $\diamond$ \textbf{GRPO\_LLaMA} & \underline{0.712} & \underline{0.281} & \underline{0.141} & \underline{0.742} & \underline{0.322} & 0.169 & \textbf{0.661} & \textbf{0.311} & \underline{0.165} & \textbf{0.689} & \textbf{0.252} & \underline{0.132} \\
\quad\quad\quad\quad $\diamond$ \textbf{GRPO\_Qwen}  & \textbf{0.734} & \textbf{0.258} & \textbf{0.118} & \textbf{0.780} & \textbf{0.251} & \textbf{0.089} & \underline{0.655} & \underline{0.319} & 0.171 & \underline{0.659} & \underline{0.265} & \textbf{0.099} \\
\bottomrule
\end{tabular}
}
\caption{Performance comparison across four states under different models and settings.}
\label{tab:hierarchical_performance}
\end{table*}

\begin{table*}
\centering

\setlength{\tabcolsep}{2pt}
\renewcommand{\arraystretch}{1.2}
\normalsize
\resizebox{\textwidth}{!}{
\begin{tabular}{l ccc ccc ccc ccc ccc}
\toprule
\multirow{2}{*}{\textbf{Model}}
& \multicolumn{3}{c}{\textbf{CA $\rightarrow$ CA}}
& \multicolumn{3}{c}{\textbf{CA $\rightarrow$ AZ}}
& \multicolumn{3}{c}{\textbf{CA $\rightarrow$ GA}}
& \multicolumn{3}{c}{\textbf{CA $\rightarrow$ OK}}
& \multicolumn{3}{c}{\textbf{AZ $\rightarrow$ CA}} \\
\cmidrule(lr){2-4} \cmidrule(lr){5-7} \cmidrule(lr){8-10} \cmidrule(lr){11-13} \cmidrule(lr){14-16}
& \textit{Acc} $\uparrow$ & \textit{Mint} $\downarrow$ & \textit{Atype} $\downarrow$
& \textit{Acc} $\uparrow$ & \textit{Mint} $\downarrow$ & \textit{Atype} $\downarrow$
& \textit{Acc} $\uparrow$ & \textit{Mint} $\downarrow$ & \textit{Atype} $\downarrow$
& \textit{Acc} $\uparrow$ & \textit{Mint} $\downarrow$ & \textit{Atype} $\downarrow$
& \textit{Acc} $\uparrow$ & \textit{Mint} $\downarrow$ & \textit{Atype} $\downarrow$ \\
\midrule
SFT\_CoPB                 & 0.743 & 0.289 & 0.154 & 0.682 & 0.420 & \underline{0.127} & 0.642 & 0.303 & 0.153 & 0.711 & 0.325 & \textbf{0.114} & 0.654 & 0.371 & \underline{0.109} \\
SFT\_RAGHome              & 0.691 & 0.349 & 0.176 & 0.671 & 0.322 & 0.183 & 0.643 & 0.324 & \underline{0.134} & 0.672 & \underline{0.246} & 0.120 & 0.636 & 0.287 & 0.151 \\
\midrule
SFT\_LLaMA            & 0.724 & \underline{0.214} & 0.110 & 0.689 & 0.256 & 0.158 & 0.632 & 0.253 & 0.195 & 0.650 & \underline{0.227} & 0.165 & 0.653 & 0.228 & 0.141 \\
SFT\_Qwen             & \underline{0.753} & 0.231 & \underline{0.104} & \underline{0.745} & \underline{0.239} & \underline{0.113} & \underline{0.708} & \underline{0.248} & 0.130 & \underline{0.714} & 0.241 & 0.127 & \underline{0.699} & \underline{0.214} & 0.097 \\
\textbf{GRPO\_LLaMA}           & 0.732 & 0.248 & 0.122 & 0.720 & 0.260 & 0.131 & 0.681 & 0.276 & 0.147 & 0.690 & 0.268 & 0.142 & 0.671 & 0.233 & 0.118 \\
\textbf{GRPO\_Qwen}            & \textbf{0.761} & \textbf{0.202} & \textbf{0.094} & \textbf{0.768} & \textbf{0.215} & \textbf{0.101} & \textbf{0.736} & \textbf{0.229} & \textbf{0.119} & \textbf{0.743} & \textbf{0.223} & \underline{0.116} & \textbf{0.721} & \textbf{0.191} & \textbf{0.086} \\
\bottomrule
\end{tabular}
}
\caption{Cross-state transfer performance. Models (SFT\_LLaMA, SFT\_Qwen, GRPO\_LLaMA, GRPO\_Qwen ) are implemented on the multi-agent architecture. CA, AZ, GA, and OK denote California, Arizona, Georgia, and Oklahoma, respectively.}
\label{tab:cross_state_transfer_california}
\end{table*}

\paragraph{Datasets.}
We conduct experiments using the 2017 National Household Travel Survey (NHTS), a nationwide travel survey covering approximately 129,696 households and 264,234 persons \cite{mcguckin2018summary}. To evaluate geographical generalization, we extract subsets from four states: California, Arizona, Georgia, and Oklahoma. Each user persona contains 12 socio-demographic features. All raw travel records are standardized into 10 semantic activity categories. Further details regarding the specific definitions of these metrics are provided in Appendix \ref{appendix:dataset}

\paragraph{Baselines Setup.}
We benchmark our approach against three categories of baselines. The first category comprises traditional deep learning models, including DeepMove \cite{feng2018deepmove} and LSTPM \cite{sun2020go}, which rely heavily on historical sequential modeling mechanisms. The second category includes prompt-based and retrieval-augmented generation frameworks, where we compare with CoPB \cite{shao2024chain} and RAGHome \cite{liu2024human}. These methods leverage advanced prompt engineering and retrieval-augmented generation to synthesize schedules. The third category evaluates zero-shot large language models. We assess representative proprietary and open-source models, including ChatGPT-5.2, Qwen3-8b, and LLaMA3.1-8b, under both single-pass generation and dual-agent architectural modes. Further details regarding the specific definitions of these metrics are provided in Appendix \ref{appendix:baseline}.

\vspace{-5pt}
\paragraph{Implementation Details.} 
We employ GPT-5.2 as the teacher model to generate Chain-of-Thought (CoT) reasoning data. Based on data, we initialize the editor agent by Supervised Fine-Tuning (SFT) on Qwen3-8B with LoRA ($r=32, \alpha=64$) for 3 epochs using a learning rate of $5\times10^{-5}$. We further optimize the SFT model with Group Relative Policy Optimization (GRPO), while keeping the same LoRA configuration. For GRPO, we train the model for 1000 steps with a learning rate of $1\times10^{-5}$, using 8 roll-out samples per prompt and a sampling temperature of 0.7. Further details regarding the specific definitions of these metrics are provided in Appendix \ref{appendix:impel}.

\paragraph{Evaluation Metrics}

\begin{table}[h]
\centering

\setlength{\tabcolsep}{4pt}
\renewcommand{\arraystretch}{1.1}
\small
% \begin{tabular}{p{0.35\linewidth}p{0.60\linewidth}}
\begin{tabular}{p{0.32\linewidth} >{\raggedright\arraybackslash}p{0.62\linewidth}}
\toprule
Category & Metrics \\
\midrule
Sequence Similarity & accuracy (\textit{Acc}), f1\_score, edit\_dist, bleu\_score \\
\midrule
Temporal Alignment & macro\_hour, micro\_hour, micro\_int, macro\_int (\textit{Mint}) \\
\midrule
Distributional Consistency & data\_jsd, act\_type (\textit{Atype}), uni\_act\_type, traj\_len \\
\bottomrule
\end{tabular}
\caption{Evaluation metrics.}
\label{tab:eval_metrics}
\vspace{-20pt}
\end{table}

\vspace{-5pt}
To evaluate generated activity chains, we use a three-level framework of 12 metrics, summarized in Table~\ref{tab:eval_metrics}. Sequence Similarity Metrics (e.g., accuracy, f1\_score, and edit\_dist) measure structural fidelity and sequential consistency with respect to reference diaries. Temporal Alignment Metrics (e.g., macro\_int and micro\_hour) assess whether generated schedules match reference timing and duration patterns at a 15-minute resolution. Distributional Consistency Metrics (e.g., data\_jsd and act\_type) use Jensen-Shannon Divergence to evaluate whether synthesized populations recover real-world aggregate regularities and semantic distributions. Detailed metric definitions are provided in Appendix~\ref{appendix:metrics}.

Table~\ref{tab:hierarchical_performance} presents the comprehensive performance comparison of all models across the four states. Beyond the absolute numerical superiority, the results reveal fundamental differences in how various architectures handle the mobility synthesis task.

\vspace{-5pt}
\subsection{Main Results}
\vspace{-5pt}

Traditional deep learning baselines, such as DeepMove and LSTPM, degrade sharply in the zero-shot setting because they rely heavily on historical trajectory sequences. Without past mobility records, they struggle to map static demographic attributes to dynamic behaviors. Prompt-based and retrieval-augmented methods, including CoPB and RAGHome, partially alleviate this issue by exploiting the language understanding of large language models, but their performance remains limited. Although prompts can steer the general structure of daily routines, they lack an explicit mechanism to enforce physical consistency, leading to spatiotemporal conflicts such as overlapping activities or incomplete 24-hour schedules.

Our fine-tuned models overcome these limitations by moving from constrained generation to active editing. In Table~\ref{tab:hierarchical_performance}, Grpo\_Qwen and Grpo\_LLaMA achieve state-of-the-art performance across all regions. In particular, Grpo\_Qwen not only delivers the highest temporal accuracy but also minimizes the Jensen-Shannon Divergence metrics (macro\_int and act\_type). These results show that our framework goes beyond generating plausible activities, enabling the synthesis of personalized, statistically accurate, and physically consistent daily schedules.

\begin{figure*}[h]
    \centering
    % --- 子图 (a): 多智能体/架构消融 ---
    \begin{subfigure}[t]{0.48\linewidth}
        \centering
        \includegraphics[width=\linewidth]{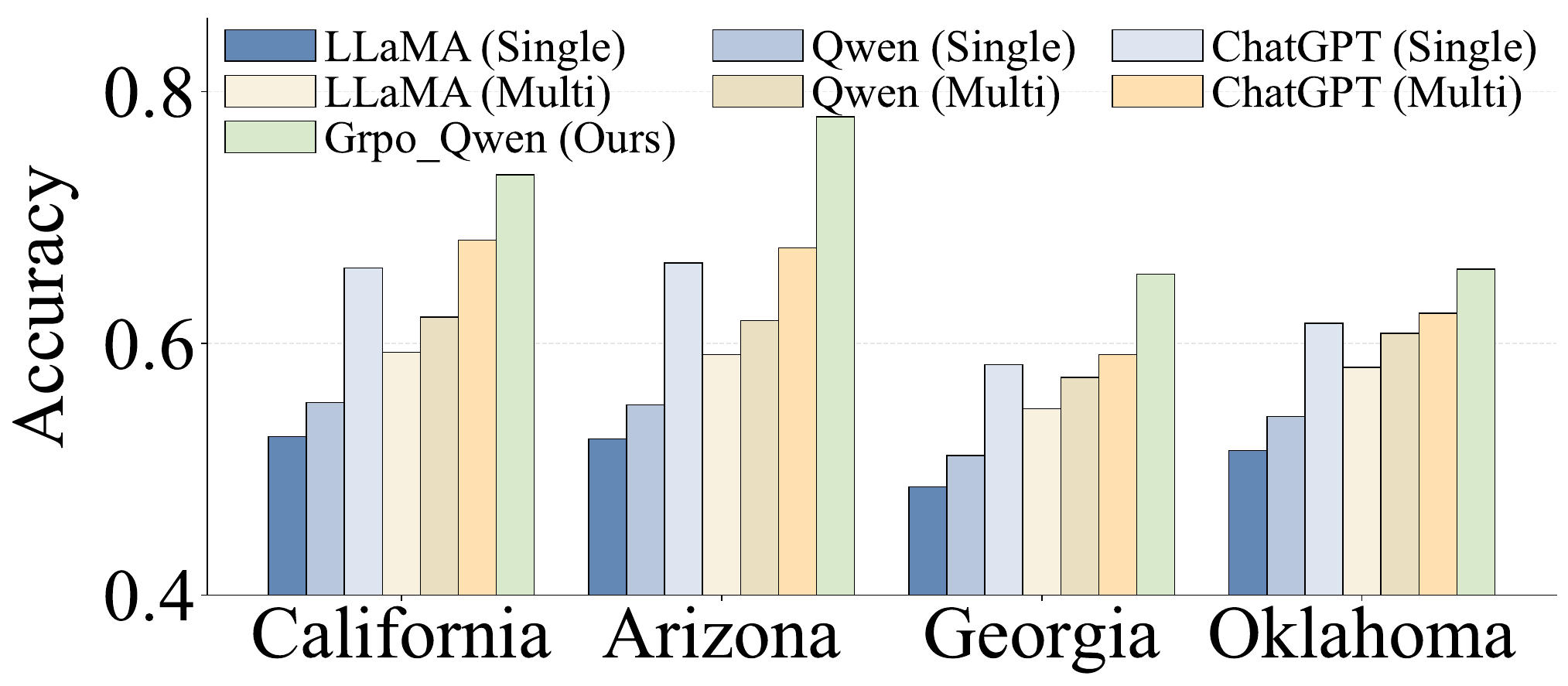}
        \caption{Single vs. Multi performance}
        \label{fig:ablation_multiagent}
    \end{subfigure}
    \hfill % 在两个子图之间插入弹性间距
    % --- 子图 (b): 训练策略 (SFT vs GRPO) 消融 ---
    \begin{subfigure}[t]{0.48\linewidth}
        \centering
        \includegraphics[width=\linewidth]{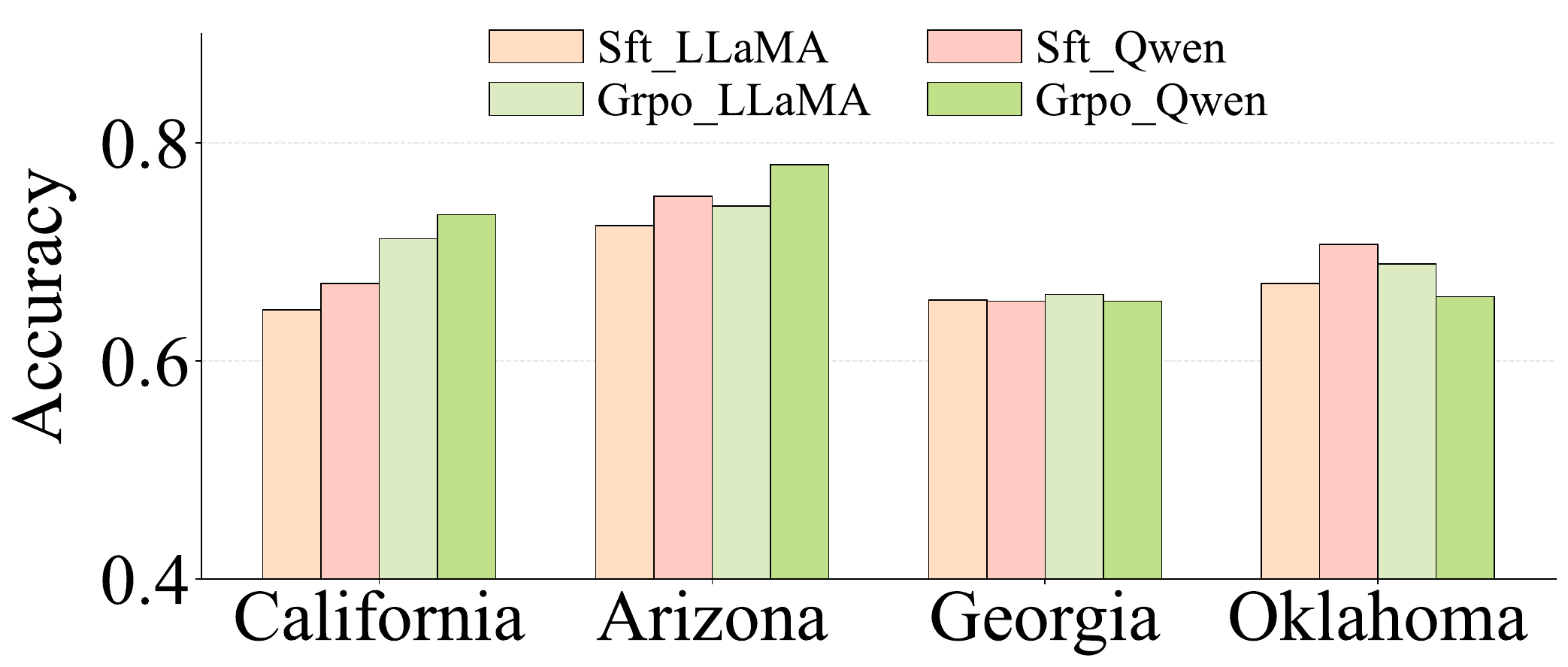}
        \caption{SFT vs. GRPO performance}
        \label{fig:ablation_sft_grpo}
    \end{subfigure}

    \caption{Ablation studies on the proposed ActivityEditor.}
    \label{fig:overall_ablation}
    \vspace{-15pt}
\end{figure*}

\vspace{-5pt}
\subsection{Zero-shot Transferability Analysis}
\vspace{-5pt}
Cold-start remains a fundamental challenge in urban mobility modeling, as unseen regions often lack sufficient historical data. To test whether our framework learns transferable mobility patterns rather than state-specific shortcuts, we conduct cross-state zero-shot transfer experiments. Models trained on California are directly test on California, Arizona, Georgia, and Oklahoma without domain adaptation, and we consider the reverse setting of Arizona $\rightarrow$ California. As shown in Table~\ref{tab:cross_state_transfer_california}, our method outperforms task-specific baselines such as CoPB~\cite{shao2024chain} and RAGHome~\cite{liu2024human}, while maintaining stable macro-level behavior statistics across target states.

These results show that our rule-guided framework captures transferable mobility logic instead of memorizing localized geographic patterns. With California as the source domain, performance drops on other states remain limited, and \texttt{Multi\_agent\_grpo\_Qwen} delivers the best overall results in most settings. Arizona $\rightarrow$ California performs worse than California $\rightarrow$ Arizona, likely because Arizona offers fewer training samples for learning transferable decision patterns. Even so, the relative ranking of model variants remains largely unchanged, demonstrating the robustness of our framework under data-scarce source domains.

\subsection{Ablation Study on Model Designs}
To disentangle the contributions of our specific architectural choices and training pipelines, we perform an in-depth ablation analysis based on the results in Table~\ref{tab:cross_state_transfer_california}.

\paragraph{Architecture Ablation on Generation Modes.} We evaluated general zero-shot large language models under both single-agent and multi-agent configurations. Across ChatGPT, Qwen, and LLaMA, the multi-agent mode consistently and noticeably outperforms the single-pass generation in Figure \ref{fig:ablation_multiagent}. This consistent performance gap highlights a crucial structural insight: daily mobility scheduling is essentially a constrained optimization problem. A single-pass autoregressive generation accumulates logical errors over a long horizon, leading to hallucinations. Introducing a multi-agent auditing mechanism provides the necessary reflection space to iteratively resolve these constraints, validating our core motivation for designing a Generate-then-Edit pipeline.

% \begin{figure}[h]
%     \centering
%     \includegraphics[width=\linewidth]{Figures/ablation_sft_grpo.pdf}
%     \caption{Caption}
%     \label{fig:ablation_sft_grpo}
% \end{figure}

\paragraph{Training Strategy Ablation on SFT and GRPO.} Figure \ref{fig:ablation_sft_grpo} and Table~\ref{tab:hierarchical_performance} show that SFT and GRPO play complementary roles in capability acquisition. SFT mainly provides statistical alignment, allowing the model to learn the mapping between demographic attributes and activity distributions and achieve competitive divergence scores. However, imitation alone is insufficient to reliably enforce hard physical constraints, leaving spatiotemporal conflicts unresolved. GRPO addresses this limitation by introducing rule-based rewards, which drive the model to internalize physical constraints and improve logical consistency during reasoning. This shift from SFT to GRPO reflects the transition from statistical fitting to constraint-aware reasoning, yielding the best overall accuracy. For the latter two states, however, the gap between SFT and GRPO becomes much smaller. This is mainly due to the extremely limited training data in these states, which makes GRPO more prone to overfitting and reduces its advantage over SFT. This result further validates the necessity and effectiveness of our Zero-shot Transferability Analysis, especially in low-resource settings.

\section{Conclusion}
We presented ActivityEditor, a dual-agent framework designed for high-fidelity, zero-shot mobility synthesis in cold-start and cross-state scenarios. By integrating intention-driven generation with rule-constrained refinement, our method captures universal mobility logic rather than merely memorizing localized patterns. Experimental results demonstrate that ActivityEditor consistently outperforms state-of-the-art baselines, achieving superior statistical fidelity and physical validity across diverse urban contexts. Future work will focus on developing a unified mobility foundation framework that integrates agentic reasoning with human mobility laws to support trajectory generation, completion, and prediction within a single modeling paradigm.

\section*{Limitations}
Despite its promising performance, several limitations of \textsc{ActivityEditor} warrant further investigation:

\begin{itemize}
    \item \textbf{Spatial Granularity:} Currently, the framework focuses on generating zone-level activity chains. It lacks the capability to synthesize precise POI-level coordinates for micro-level urban analysis. Future research will explore the integration of external POI-retrieval tools and urban perception modules to assist the LLM in achieving more fine-grained spatial generation.
    
    \item \textbf{Computational Efficiency:} The multi-turn iterative refinement between the Intention-driven and Editor agents incurs significant computational overhead, which may limit its scalability for massive population simulations. Knowledge distillation into smaller language models or the optimization of reinforcement learning efficiency represents a viable direction to reduce inference latency.
    
    \item \textbf{Robustness and Hallucination:} As an LLM-based framework, the output of \textsc{ActivityEditor} is not fully controllable. While we utilize a structured parser to extract the expected context, it may occasionally fail. Furthermore, due to the inherent potential for hallucination, the agent may generate "phantom" addresses that do not exist in the real world. While verifying this against a valid location list is feasible in controlled experiments, ensuring absolute robustness in open-world urban applications remains a significant challenge.
\end{itemize}

\section*{Ethical Considerations}
All human mobility data used in the experiments come
from publicly available open-source datasets\cite{mcguckin2018summary}. We do not attempt to extract any personal information from
these datasets.

\section*{AI Usage}

We state that AI-assisted technologies were used strictly for the purpose of language polishing and grammar correction to enhance the clarity of our writing. The authors take full responsibility for the integrity of the work, including the validity of all experimental results and the accuracy of all references.

\newpage
\bibliography{ref}

\newpage
\appendix
\section{Appendix}
\subsection{Dataset Description and Analysis}
\label{appendix:dataset}
\begin{figure*}[h] 
\centering 
\includegraphics[width=1\textwidth]{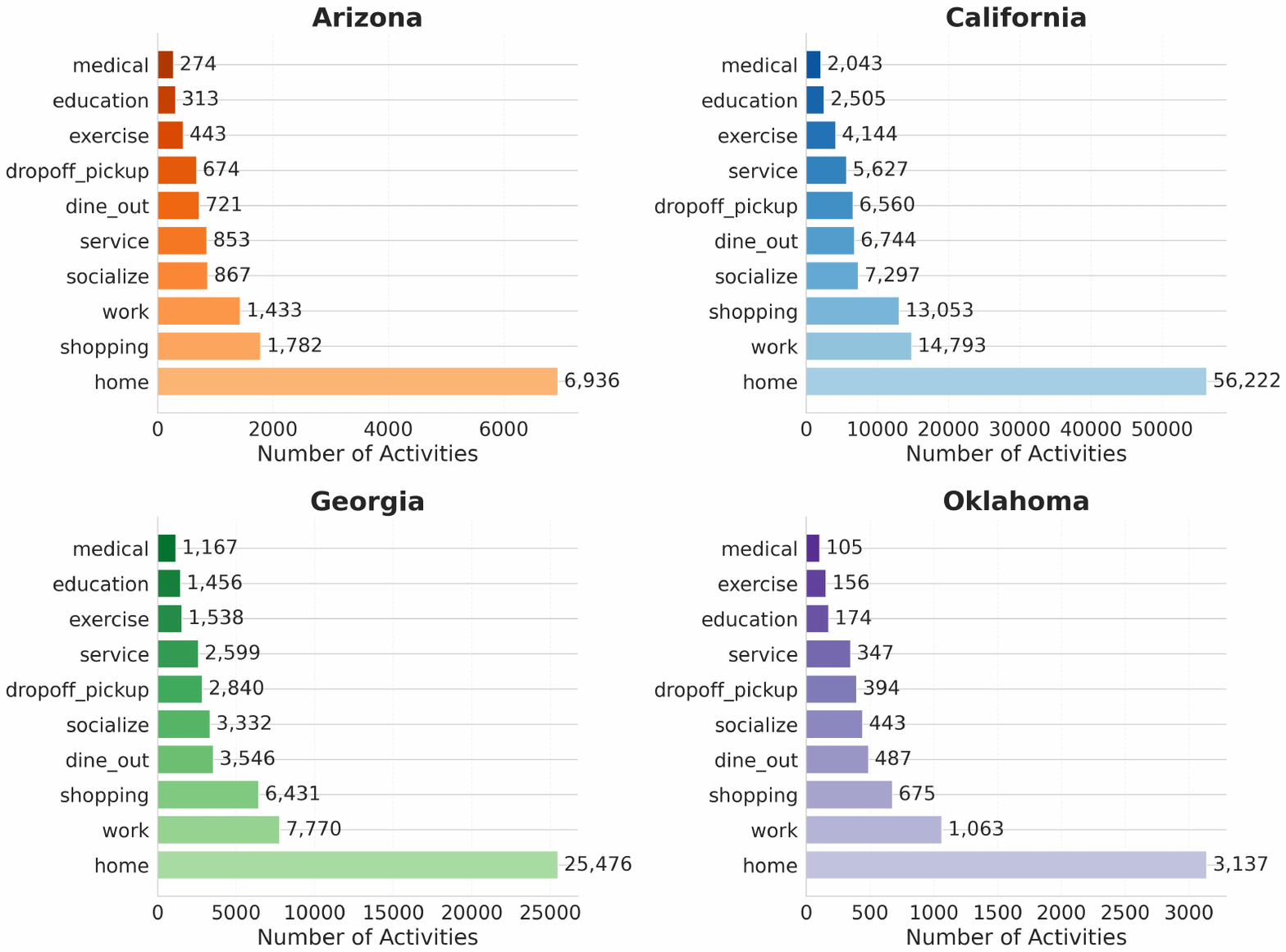} 
\caption{Activity number distribution by state.} 
\label{fig:activity_number} 
\vspace{-0.5em} 
\end{figure*}

To comprehensively evaluate our proposed framework, we utilize the 2017 National Household Travel Survey (NHTS) dataset. This dataset provides rich socio-demographic profile information paired with highly detailed 24-hour activity and travel diaries for residents across the United States. We specifically extract subsets from four representative states: California, Arizona, Georgia, and Oklahoma. These states encompass high diversity in urban structures, economic development levels, and population densities. As illustrated in Figure \ref{fig:activity_number}, the volume of available trajectory data varies significantly across different regions. California possesses over 56,000 home activity records, followed by Georgia with over 25,000, Arizona with approximately 6,900, and Oklahoma with roughly 3,100. This severe data imbalance highlights a critical real-world challenge: many newly planned or smaller regions lack sufficient historical data to support the training of reliable predictive models. This discrepancy directly motivates our zero-shot cross-regional generation task, which aims to transfer mobility knowledge from data-rich source domains to data-scarce target domains.

\begin{figure*}[h] 
\centering 
\includegraphics[width=1\textwidth]{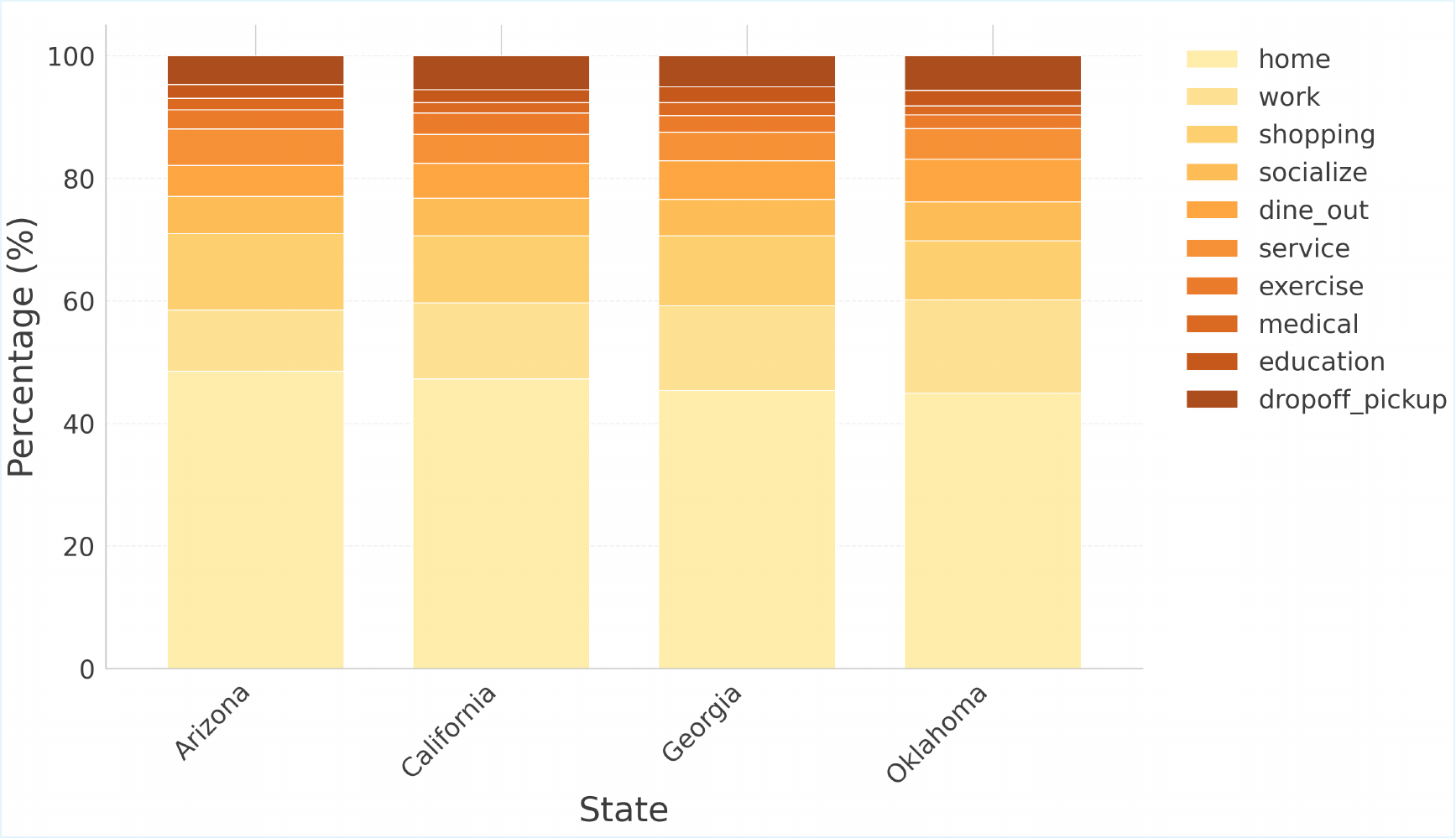} 
\caption{Activity type distribution by state.} 
\label{fig:activity_type} 
\vspace{-0.5em} 
\end{figure*}

\begin{figure*}[h] 
\centering 
\includegraphics[width=1\textwidth]{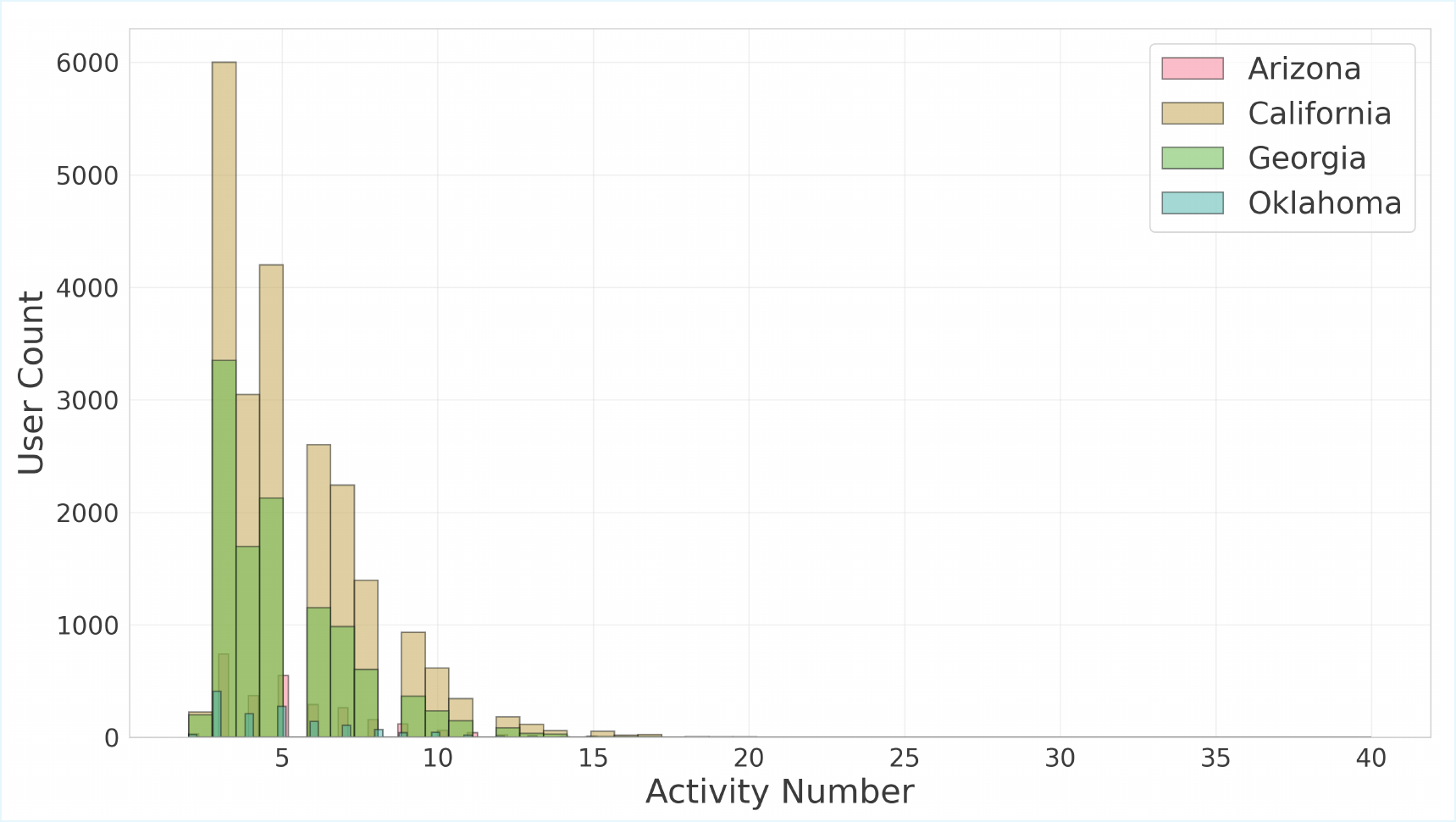} 
\caption{Chain length distribution by state.} 
\label{fig:chain_length} 
\vspace{-0.5em} 
\end{figure*}

Despite the vast differences in data scale and geospatial layouts, the macroscopic statistical features of the dataset reveal that fundamental human mobility patterns remain highly consistent, which strongly validates the theoretical feasibility of our transfer learning approach. Figure \ref{fig:activity_type} presents the stacked percentage distribution of activity types across the four states. The proportion of human activities remains universally stable across different regions, with home, work, and shopping consistently dominating daily routines. Similarly, Figure \ref{fig:chain_length} demonstrates that the distribution of daily activity chain lengths follows a nearly identical right-skewed pattern across all states, typically peaking between three to five activities per person. This cross-regional consistency provides empirical support for our core assumption: human mobility intentions are fundamentally driven by socio-demographic attributes rather than determined by local spatial geometric features. It is exactly this ubiquitous similarity that enables our intention-driven agent to successfully generalize to unseen cities in a zero-shot manner.

\begin{figure*}[h] 
\centering 
\includegraphics[width=1\textwidth]{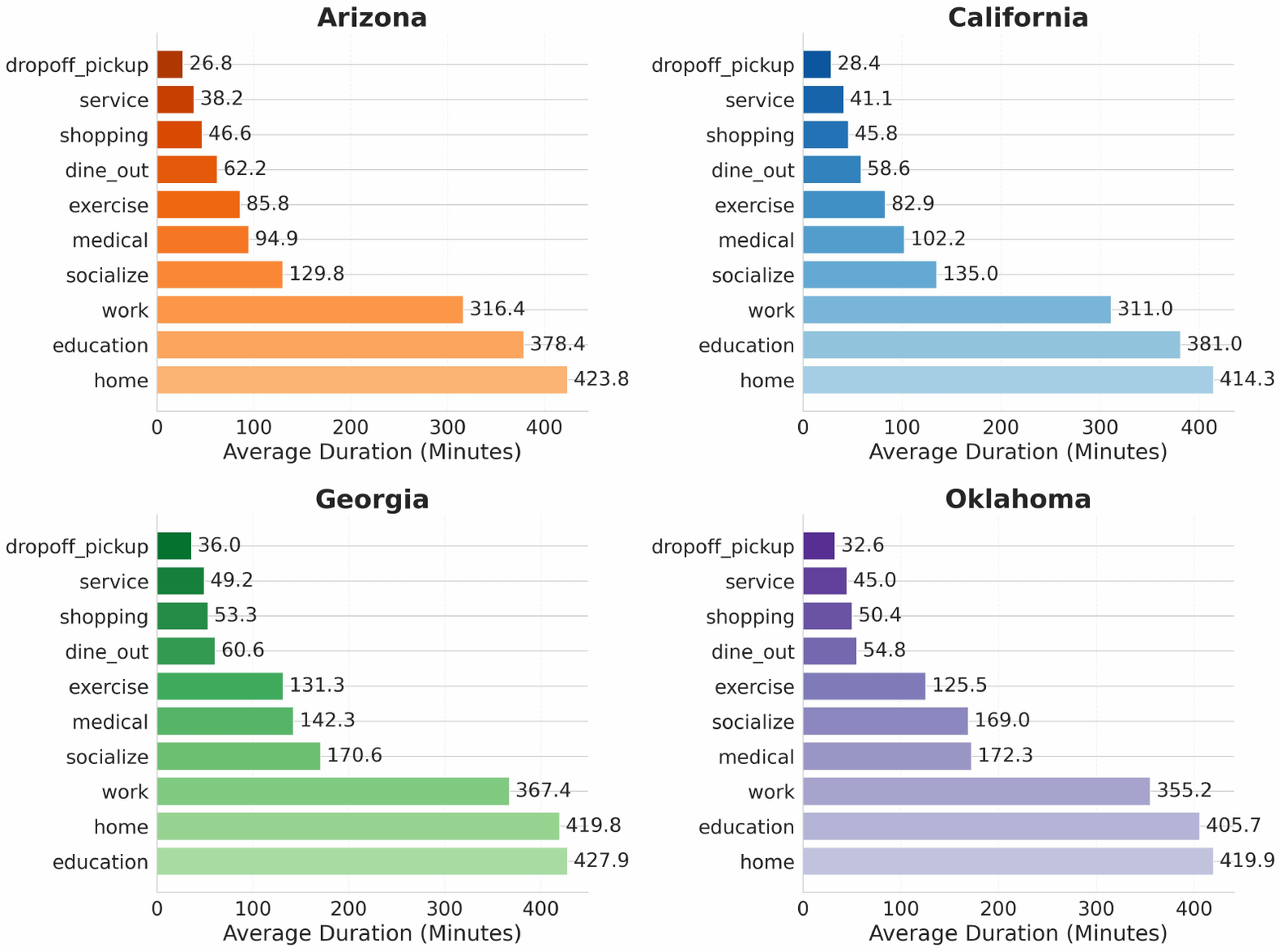} 
\caption{Activity duration by state.} 
\label{fig:activity_duration} 
\vspace{-0.5em} 
\end{figure*}

Furthermore, the dataset clearly reflects the strict physical boundaries and physiological constraints governing human behavior. Figure \ref{fig:activity_duration} visualizes the average duration of various activities across different states. The temporal patterns strictly adhere to human commonsense: rigid activities such as sleeping at home or working typically span hundreds of minutes, whereas transitional or service activities, such as dropping off children, generally remain stable under forty minutes. These universal temporal regularities do not fluctuate across state borders. This observation fundamentally underpins our design of the rule-based reward model in the Group Relative Policy Optimization phase. By transforming these ubiquitous physiological and temporal constraints into programmatic reward signals, the editor agent is able to autonomously internalize the authentic logic of mobility behavior, thereby eliminating logical hallucinations without requiring any target-domain training data.

\subsection{Detailed Description of Baselines}
\label{appendix:baseline}

To comprehensively evaluate the performance of the proposed framework, four representative baseline methods are selected for comparison. These methods are divided into two categories: traditional deep learning baselines and large language model baselines.

\textbf{DeepMove} \cite{feng2018deepmove}:  
DeepMove is a classic human mobility prediction model. It employs an attentional recurrent neural network to model sequential transition patterns and multi-level periodic behaviors in users' historical trajectories, thereby enabling the prediction of future mobility behavior.

\textbf{LSTPM} \cite{sun2020go}:  
LSTPM is designed to capture both long-term and short-term user mobility preferences. By combining a context-aware non-local network with a geo-dilated recurrent neural network, it effectively models users' long-term preferences and short-term dynamic changes in point-of-interest visits.

\textbf{CoPB} \cite{shao2024chain}:  
CoPB is grounded in the psychological Theory of Planned Behavior and uses a structured prompt engineering workflow to guide large language models to first reason about users' subjective beliefs and behavioral intentions, and then generate daily activity sequences, thereby improving the logical consistency and plausibility of mobility generation.

\textbf{RAGHome} \cite{liu2024human}:  
RAGHome introduces a retrieval-augmented generation mechanism that retrieves historical mobility records of individuals with characteristics similar to the target user and incorporates them as contextual examples into the large language model, thereby enhancing generation quality and realism in complex spatiotemporal tasks.

\subsection{Prompt Examples}
\label{app:prompt_examples}

For completeness, this subsection summarizes the core prompts used throughout our trajectory generation framework.

\subsubsection{Prompt of Intention-driven Agent}

\begin{lstlisting}[style=promptstyle]
You are a daily schedule planner. Given a person's profile, generate a complete and realistic 24-hour schedule with exact start/end times.

RULES:
- Use ONLY these activity types: home, work, education, shopping, service, medical, dine_out, socialize, exercise, dropoff_pickup
- Schedule MUST cover full 24 hours: start at 00:00, end at 24:00, NO gaps or overlaps
- Must start and end with "home"
- Consecutive activities must connect (end_time of one = start_time of the next)
- Durations must be realistic based on the person's profile

OUTPUT FORMAT:
{
  "reasoning": "Brief explanation of the schedule choices",
  "schedule": [
    {"activity": "home", "start_time": "00:00", "end_time": "07:30"},
    {"activity": "work", "start_time": "07:30", "end_time": "17:00"},
    {"activity": "home", "start_time": "17:00", "end_time": "24:00"}
  ]
}

USER MESSAGE FORMAT:
Generate a complete 24-hour schedule for this person:

- Age range: {age_range}
- Gender: {gender}
- Race: {race}
- Education: {education}
- Employment status: {employment_status}
- Work schedule: {work_schedule}
- Occupation: {occupation}
- Primary activity: {primary_activity}
- Work from home: {work_from_home}
- Driver on travel day: {driver_on_travel_day}
- Distance to work (miles): {distance_to_work_miles}
- Work state: {work_state}
\end{lstlisting}

\subsubsection{Prompt of Editor Agent}

\begin{lstlisting}[style=promptstyle]
You are an Editor Agent for daily schedule refinement.

Your task:
1. Check the INITIAL SCHEDULE against 5 constraint types:
   Physical, Logical, Common Sense, Temporal, and Coherence
2. Identify any violations in the draft
3. Apply edit operations from the action space:
   ADD / DELETE / SHIFT / REPLACE / SPLIT
4. Output the final constraint-satisfying schedule

OUTPUT FORMAT:
[THOUGHT]
Constraint Checking:
1. Physical Constraints (Hard): overlaps, 24h coverage, ends at 24:00
2. Logical Constraints (Hard): starts/ends at home, starts at 00:00, consecutive identical activities must be merged
3. Common Sense Constraints (Soft): activities match socio-demographic profile
4. Temporal Constraints (Soft): realistic durations for each activity type
5. Coherence Constraints (Soft): logical transitions, not over-fragmented

Edit Operations Applied:
- ADD: activity 'Y' at time HH:MM (reason)
- DELETE: activity 'X' at index N (reason)
- SHIFT: activity 'Z' start/end time adjustment (reason)
- REPLACE: activity 'A' -> 'B' (reason)
- SPLIT: activity 'X' divided into two segments (reason)

Final Result: All constraints satisfied / Violations remain
[/THOUGHT]

[JSON]
[final schedule as JSON array]
[/JSON]

USER MESSAGE FORMAT:
PERSON PROFILE:
{user_profile as JSON}

INITIAL SCHEDULE:
{initial_schedule as JSON}
\end{lstlisting}

\subsubsection{Prompt of Data Generation Teacher Model}

\begin{lstlisting}[style=promptstyle]
SYSTEM PROMPT:
You are a schedule planning expert. You understand human behavior patterns and generate realistic daily schedules.

USER MESSAGE FORMAT:
You are a Critic and Editor Agent. Your job is to validate and refine a daily schedule.

PERSON PROFILE:
{user_profile as JSON}

INITIAL SCHEDULE (from unified 2-stage generator):
{initial_schedule as JSON}

GROUND TRUTH REFERENCE (for comparison):
{ground_truth_schedule as JSON}

YOUR TASK:
Act as the Editor. Check constraints on the INITIAL SCHEDULE, identify violations, and apply edits to match the GROUND TRUTH.

OUTPUT FORMAT:
[THOUGHT]
Constraint Checking:
1. Physical (Hard): overlaps? 24h coverage? ends at 24:00? -> Yes / No
2. Logical (Hard): starts/ends home? starts at 00:00? -> Yes / No
3. Common Sense (Soft): activities match profile? -> Yes / No
4. Temporal (Soft): realistic durations? -> Yes / No
5. Coherence (Soft): logical flow? not fragmented? -> Yes / No

Edits to Match Ground Truth:
If already matches: No edits needed
Otherwise list each operation:
- DELETE: 'X' at idx N (reason)
- ADD: 'Y' at time HH:MM (reason)
- SHIFT: 'Z' time adjustment (reason)
- REPLACE: 'A' -> 'B' (reason)

Final Result:
All constraints satisfied after edits? Yes / No
[/THOUGHT]

[JSON]
[final schedule - must match the GROUND TRUTH]
[/JSON]

Be thorough in checking each constraint on the INITIAL SCHEDULE.
Show edit operations explicitly if violations are found.
\end{lstlisting}

\subsubsection{Prompt of SFT Student Model}
\label{appendix:prompts}

\begin{lstlisting}[style=promptstyle]
You are a Critic and Editor Agent for daily schedule refinement.

Your task:
1. Check the INITIAL SCHEDULE against 5 constraint types:
   Physical, Logical, Common Sense, Temporal, and Coherence
2. Identify any violations
3. Apply edit operations:
   ADD / DELETE / SHIFT / REPLACE
4. Output the refined schedule

OUTPUT FORMAT:
[THOUGHT]
Constraint Checking:
1. Physical Constraints (Hard): overlaps, 24h coverage, ends at 24:00
2. Logical Constraints (Hard): starts/ends at home, starts at 00:00
3. Common Sense Constraints (Soft): age/employment appropriate activities
4. Temporal Constraints (Soft): realistic durations
5. Coherence Constraints (Soft): logical transitions, not over-fragmented

Applying Edit Operations:
- DELETE: activity 'X' at index N (reason)
- ADD: activity 'Y' at time HH:MM (reason)
- SHIFT: activity time adjustment (reason)
- REPLACE: activity type change (reason)

Final Result: Yes / No
[/THOUGHT]

[JSON]
[refined schedule as JSON array]
[/JSON]

USER MESSAGE FORMAT:
PERSON PROFILE:
{user_profile as JSON}

INITIAL SCHEDULE:
{initial_schedule as JSON}

TRAINING TARGET:
[THOUGHT]
{teacher-generated constraint-checking trace and edit operations}
[/THOUGHT]

[JSON]
{final refined schedule matching ground truth}
[/JSON]
\end{lstlisting}

\subsubsection{Prompt of GRPO Training}

\begin{lstlisting}[style=promptstyle]
You are a daily activity schedule generator. Your task is to generate a realistic daily schedule for a person based on their profile.

OUTPUT FORMAT:
[THOUGHT]
Brief reasoning about the person's schedule patterns.
[/THOUGHT]

[JSON]
[
  {"activity": "home", "start_time": "00:00", "end_time": "07:00"}
]
[/JSON]

The schedule must start at 00:00 and end at 24:00, covering the full day without gaps or overlaps.

USER MESSAGE FORMAT:
Generate a daily schedule for this person:
{user_profile as JSON}
\end{lstlisting}

\subsection{Detailed Parameter Settings}
\label{appendix:impel}
To ensure the reproducibility of our experiments, we provide the complete hyperparameter configurations for both the Supervised Fine-Tuning (SFT) and Group Relative Policy Optimization (GRPO) stages in Table 4. 

\begin{table}[h]
\centering

\small
\begin{tabular}{lc}
\toprule
Hyperparameter & SFT Stage \\
\midrule
Base Model & Qwen3-8B \\
LoRA Rank (r) & 32 \\
LoRA Alpha & 64 \\
LoRA Dropout & 0.05 \\
Training Epochs & 3 epochs \\
Peak Learning Rate & 5e-5 \\
Learning Rate Scheduler & Cosine \\
Warmup & 50 steps \\
Per-device Batch Size & 2 \\
Gradient Accumulation Steps & 8 \\
Max Sequence Length & 4096 \\
\bottomrule
\end{tabular}
\caption{Hyperparameter settings for the SFT stage.}
\label{tab:hyperparameters_sft}
\end{table}

\begin{table}[h]
\centering

\small
\begin{tabular}{lc}
\toprule
Hyperparameter & GRPO Stage \\
\midrule
Base Model & Qwen3-8B \\
LoRA Rank (r) & 32 \\
LoRA Alpha & 64 \\
LoRA Dropout & 0.05 \\
Training Steps & 1000 steps \\
Peak Learning Rate & 1e-5 \\
Learning Rate Scheduler & Cosine \\
Warmup & 0.1 ratio \\
Gradient Accumulation Steps & 4 \\
Max Sequence Length & 4096\\
Roll-out Samples per Prompt & 8 \\
Sampling Temperature & 0.7 \\
Min\_p & 0.1 \\
\bottomrule
\end{tabular}
\caption{Hyperparameter settings for the GRPO stage.}
\label{tab:hyperparameters_grpo}
\end{table}

All models were trained using the bfloat16 precision format to optimize memory usage while maintaining numerical stability. The AdamW optimizer was utilized across all training phases, with the beta1 and beta2 parameters set to 0.9 and 0.99, respectively, alongside a weight decay coefficient of 0.1. To prevent gradient explosion, a gradient clipping threshold of 1.0 was universally applied across all parameter updates. As stated in the main text, both stages utilized the exact same Low-Rank Adaptation configuration to efficiently update the attention and feed-forward modules. The specific trainable adapters were applied to the q\_proj, k\_proj, v\_proj, o\_proj, gate\_proj, up\_proj, and down\_proj layers. The dataset was split into training, validation, and testing sets with a ratio of 8:1:1, and all reported results were averaged over three independent runs.

\subsection{Evaluation Metrics Formulation}
\label{appendix:metrics}
Before computing the metrics, each continuous daily schedule is discretized into a fixed-length sequence of 96 time slots, representing 15-minute intervals. The assigned activity for each slot is the one with the longest temporal overlap. Let $G \in \mathbb{Z}^{N \times 96}$ and $T \in \mathbb{Z}^{N \times 96}$ denote the generated and ground truth slot matrices for $N$ users, respectively. We group the 12 metrics into three mathematical categories.

\paragraph{Sequence Similarity Metrics.}
These metrics evaluate how closely the generated activity sequence matches the ground-truth sequence at the token and structural levels.

Accuracy (Acc) calculates the slot-level exact match rate between the generated and ground truth sequences:
\begin{equation}
\text{Acc} = \frac{1}{N \times 96} \sum_{i=1}^{N}\sum_{t=1}^{96} \mathbf{1}[G_{i,t} = T_{i,t}]
\end{equation}

Macro F1-Score computes the balanced accuracy over $C=11$ activity classes:
\begin{equation}
\text{F1}_{\text{macro}} = \frac{1}{K} \sum_{k=1}^{K} \frac{2 \cdot \text{Prec}_k \cdot \text{Recall}_k}{\text{Prec}_k + \text{Recall}_k}
\end{equation}

Edit Distance (edit\_dist) calculates the minimum edit cost, normalized by the sequence length:
\begin{equation}
\text{edit\_dist} = \frac{1}{N} \sum_{i=1}^{N} \frac{\text{Levenshtein}(G_i, T_i)}{96}
\end{equation}

BLEU Score applies n-gram sequence overlap precision up to 4-gram:
\begin{equation}
\text{BLEU} = \frac{1}{N} \sum_{i=1}^{N} \text{sentence\_BLEU}(T_i, G_i)
\end{equation}

\paragraph{Temporal Alignment Metrics.}
These metrics measure whether the generated schedules preserve realistic temporal patterns, including activity durations and starting times.

These metrics utilize the Jensen-Shannon Divergence (JSD) to evaluate duration and start-time distributions.

Macro\_int evaluates the marginal distribution of continuous activity durations (run lengths), denoted as frequency vector $m$:
\begin{equation}
\text{macro\_int} = \text{JSD}\left(\frac{m^{\text{gt}}}{\|m^{\text{gt}}\|_1} \bigg\| \frac{m^{\text{gen}}}{\|m^{\text{gen}}\|_1}\right)
\end{equation}

Micro\_int computes the divergence on the flattened joint distribution matrix $M$ of activity types and durations:
\begin{equation}
\text{micro\_int} = \text{JSD}\left(\frac{M^{\text{gt}}_{\text{flat}}}{\|M^{\text{gt}}\|_1} \bigg\| \frac{M^{\text{gen}}_{\text{flat}}}{\|M^{\text{gen}}\|_1}\right)
\end{equation}

\begin{table*}   % 改这里，[H] → [t]
\centering
\small

\begin{tabular}{p{2.8cm}p{3.8cm}p{2.8cm}p{2.8cm}}
\toprule
\textbf{Initial Draft} & \textbf{Reasoning Cues} &
\textbf{Edited Output} & \textbf{Ground Truth} \\
\midrule
{[00:00--07:45] Home} &  & {[00:00--07:45] Home} &
{[00:00--07:45] Home} \\
{[07:45--16:45] Work} & occupation $\rightarrow$ Evening
Work & {[07:45--16:30] Work} & {[07:45--16:30] Work} \\
{[16:45--18:00] Home} & relationship $\rightarrow$
Shopping & {[16:30--17:20] Shopping} & {[16:30--17:30]
Shopping} \\
{[17:45--19:00] Work} & constraint $\rightarrow$ Fix
overlap & {[17:20--19:50] Work} & {[17:30--17:45] Service}
\\
{[19:00--19:35] Shopping} & short commute $\rightarrow$ no
extra Home stay & {[19:50--24:00] Home} & {[17:45--20:30]
Work} \\
{[19:35--24:00] Home} &  &  & {[20:30--24:00] Home} \\
\midrule
\textbf{Status:} invalid & \textbf{Signals:} profile +
constraint check & \textbf{Status:} valid &
\textbf{Status:} observed \\
\bottomrule
\end{tabular}
\caption{Case study of profile-grounded trajectory
editing.}
\label{tab:case_study_comparison}
\end{table*}

Macro\_hour measures the divergence of the marginal distribution of activity onset times (start slots), denoted as vector $h$:
\begin{equation}
\text{macro\_hour} = \text{JSD}\left(\frac{h^{\text{gt}}}{\|h^{\text{gt}}\|_1} \bigg\| \frac{h^{\text{gen}}}{\|h^{\text{gen}}\|_1}\right)
\end{equation}

Micro\_hour calculates the divergence on the joint distribution matrix $H$ of activity types and their onset times:
\begin{equation}
\text{micro\_hour} = \text{JSD}\left(\frac{H^{\text{gt}}_{\text{flat}}}{\|H^{\text{gt}}\|_1} \bigg\| \frac{H^{\text{gen}}_{\text{flat}}}{\|H^{\text{gen}}\|_1}\right)
\end{equation}

\paragraph{Distributional Consistency Metrics.}
These metrics assess whether the generated population reproduces the global statistical properties and behavioral diversity of the real-world data.

Data\_jsd evaluates population-level trajectory diversity by treating each full 96-slot sequence as a single token with probability $p$:
\begin{equation}
\text{data\_jsd} = \text{JSD}(p^{\text{gt}} \| p^{\text{gen}})
\end{equation}

Act\_type calculates the divergence of the total time budget allocated to each activity category $p_k$:
\begin{equation}
\text{act\_type} = \text{JSD}(p_k^{\text{gt}} \| p_k^{\text{gen}})
\end{equation}

Uni\_act\_type computes the divergence based on the frequency of distinct activity episodes per category $q_k$:
\begin{equation}
\text{uni\_act\_type} = \text{JSD}(q_k^{\text{gt}} \| q_k^{\text{gen}})
\end{equation}

Traj\_len measures the divergence in the distribution of the total number of distinct activity episodes per user, denoted as count vector $r$:
\begin{equation}
\text{traj\_len} = \text{JSD}\left(\frac{r^{\text{gt}}}{\|r^{\text{gt}}\|_1} \bigg\| \frac{r^{\text{gen}}}{\|r^{\text{gen}}\|_1}\right)
\end{equation}

\subsection{Case Study: Profile-Grounded Trajectory Editing}

We present a case study based on a real user from the NHTS California subset. The profile indicates that the user isrelationship a female, a mother, employed full-time in a professional, managerial, or technical occupation, does not work from home, drives on the travel day, and has a short commute distance of 4.15 miles. Her primary activity is working.

\paragraph{Reasoning trace.}
Reasoning process is as follows:

\begin{quote}
\texttt{[THOUGHT]}\\
\textbf{Constraint Checking (5 types):}

1. \textbf{Physical Constraints (Hard):}
- Detected overlap: Home 16:45--18:00 and Work 17:45--19:00\\
\textbf{Result: FAIL}

2. \textbf{Logical Constraints (Hard):}
- Starts at Home? Yes
- Ends at Home? Yes
- Full-day coverage? Yes\\
\textbf{Result: PASS}

3. \textbf{Common Sense Constraints (Soft):}
- relationship $\rightarrow$ Shopping
- occupation $\rightarrow$ Evening Work\\
\textbf{Result: informative cues}

4. \textbf{Temporal Constraints (Soft):}
- Shopping is more plausible after daytime Work than after evening Work\\
\textbf{Result: needs adjustment}

5. \textbf{Coherence Constraints (Soft):}
- short commute $\rightarrow$ avoid extra Home stay before returning to Work\\
\textbf{Result: needs adjustment}

\textbf{Editing decision:}
- shift the first Work block earlier,
- move Shopping to the interval after daytime Work,
- delay the second Work block until after Shopping,
- remove the redundant late Shopping segment.

\textbf{Output:}
- Home 00:00--07:45
- Work 07:45--16:30
- Shopping 16:30--17:20
- Work 17:20--19:50
- Home 19:50--24:00

\textbf{Result:}
valid schedule with improved profile consistency and closer structural alignment to the observed trajectory.
\texttt{[/THOUGHT]}
\end{quote}

This case study shows the role of the Editor Agent: improving a profile-informed draft by fixing hard errors and refining activity order with profile-conditioned commonsense. The result is not an exact match to the observed day, but a more realistic and coherent mobility trajectory.

\label{sec:appendix}

%1.dataset description:数据集描述 可视化 多放点
%2.baseline的具体描述 baseline1:具体的方法
%3.完整的实验表格 
%4.prompt
%5.case study

\end{document}